\documentclass{IEEEtran}
\usepackage{times}
\usepackage{comment}
\usepackage{epsfig}
\usepackage{graphicx}
\usepackage{amsmath}
\usepackage{amssymb}
\usepackage{graphicx}
\usepackage{url}
\usepackage{array}
\usepackage{multirow}
\usepackage{subcaption}
\usepackage{color}
\usepackage{epstopdf}

\usepackage{fancyhdr}
\fancypagestyle{firstpage}{%
  \rhead{Published in ICARCV 2018. Final paper is available at: https://ieeexplore.ieee.org/document/8581147.}
  \lfoot{\small{\copyright2018 IEEE. Personal use of this material is permitted. Permission from IEEE must be obtained for all other users, including reprinting/republishing this material for advertising or promotional purposes, creating new collective works for resale or redistribution to servers or lists, or reuse of any copyrighted components of this work in other works.}}
}

\title{RCCNet: An Efficient Convolutional Neural Network for Histological Routine Colon Cancer Nuclei Classification
}

\author{S H Shabbeer Basha, Soumen Ghosh, Kancharagunta Kishan Babu, Shiv Ram Dubey, \\ Viswanath Pulabaigari, and Snehasis Mukherjee
\thanks{Shabbeer Basha S.H., S. Ghosh, K.K. Babu, S.R. Dubey, Viswanath P., S. Mukherjee  are with Computer Vision and Machine Learning Groups, Indian Institute of Information Technology, Sri City, Andhra Pradesh - 517646, India. 
{\tt\small email: \{shabbeer.sh, soumen.g, kishanbabu.k, srdubey, viswanath.p, snehasis.mukherjee\}@iiits.in}}%
}

\begin{document}
\maketitle
\thispagestyle{firstpage}

\begin{abstract}

Efficient and precise classification of histological cell nuclei is of utmost importance due to its potential applications in the field of medical image analysis. It would facilitate the medical practitioners to better understand and explore various factors for cancer treatment. The classification of histological cell nuclei is a challenging task due to the cellular heterogeneity. This paper proposes an efficient Convolutional Neural Network (CNN) based architecture for classification of histological routine colon cancer nuclei named as RCCNet. The main objective of this network is to keep the CNN model as simple as possible. The proposed RCCNet model consists of $1,512,868$ learnable parameters which are significantly less compared to the popular CNN models such as AlexNet, CIFAR-VGG, GoogLeNet, and WRN. The experiments are conducted over publicly available routine colon cancer histological dataset ``CRCHistoPhenotypes". The results of the proposed RCCNet model are compared with five state-of-the-art CNN models in terms of the accuracy, weighted average F1 score and training time. The proposed method has achieved a classification accuracy of $80.61\%$ and $0.7887$ weighted average F1 score. The proposed RCCNet is more efficient and generalized in terms of the training time and data over-fitting, respectively.
\end{abstract}

\section{Introduction}

The medical image analysis is one of the fundamental, applied, and active research area during the last few decades. The classification of medical images such as colon cancer is one of the most popular core research areas of the medical image analysis~\cite{sirinukunwattana2016locality}. Categorization of tumors at the cellular level can help medical professionals to better understand the tumor characteristics which can facilitate them to explore various options for cancer treatment. Classifying cell nuclei from routine colon cancer (RCC) images is a challenging task due to cellular heterogeneity.

The American Cancer Society publishes colon cancer (also known as Colorectal cancer (CRC)) statistics every three years. The American Cancer Society Colorectal Cancer Facts \& Figures 2017-2019~\cite{siegel2017colorectal} reports the following. 
In $2017$, in the USA, an estimation says that $95,520$ new cases of colon cancer were found out of which $50,260$ people died, which includes $27,150$ men and $23,110$ women. The colon cancer  is the third most dangerous cancer which affects both men and women. Thus, it is required to analyze the medical images for accurate colon cancer disease recognition. 

Nucleus image classification has been applied to various histology related medical applications. 
Following are some recent attempts in applying image analysis or computer vision techniques in the medical domain. 
In 2014, Veta et al.~\cite{veta2014breast} published a complete review article on breast cancer image analysis. Many researchers worked in the area of histological image analysis, a few of them are~\cite{arif2007classification, sharma2015multi}.  Traditional machine learning methods have been employed by several researchers using handcrafted features obtained from histology images~\cite{jones2009scoring, chang2013classification}. Manually engineered features may not always represent the underlying structure of histology images. On the other hand, convolutional neural networks (CNNs) extract high-level and more semantic features automatically from the training data. 

\begin{figure*}
\includegraphics[width=\linewidth, height = 4 cm]{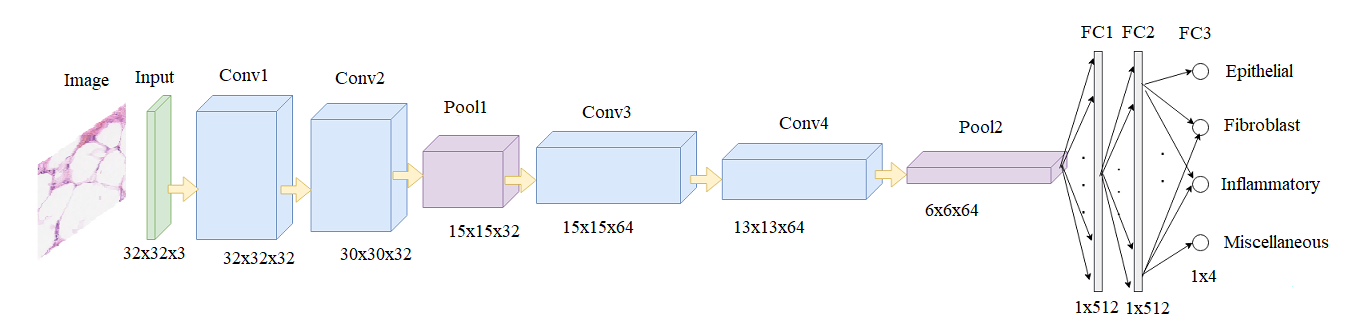}
\setlength{\abovecaptionskip}{-11pt}
\setlength{\belowcaptionskip}{-15pt}
  \caption{Proposed RCCNet architecture for routine colon cancer nuclei classification. The first two convolutional layers (i.e., $Conv1$ and $Conv2$ layers) have $32$ filters of size $3\times3$. The two convolutional layers in the middle (i.e., $Conv3$ and $Conv4$ layers) have $64$ filters of size $3\times3$, each. The last three fully connected layers (i.e., $FC1$, $FC2$ and $FC3$ layers) have $512$, $512$, and $4$ neurons, respectively. Two pooling layers (i.e., $Pool1$ and $Pool2$ layers) are used to down-sample the spatial dimension after $Conv2$ and $Conv4$ layers, respectively.}

\label{model}
\end{figure*}

Recently, deep learning based approaches have achieved very promising performance in the field of computer vision and image analysis~\cite{lecun2015deep}. In $2012$, Krizhevsky et al.~\cite{krizhevsky2012imagenet} proposed a deep CNN model (called the AlexNet) consisting of 8 learnable layers for image classification. The AlexNet model is further extended to VGG-16 by Simonyan et al.~\cite{simonyan2014very} with $16$ number of trainable layers. Later, the GoogLeNet with inception modules became popular for deep networks~\cite{szegedy2015going}. In recent development, He et al.~\cite{he2016deep} proposed a deeper residual network (ResNet) with $152$ layers for image recognition in $2016$. The CNN based models have also shown very encouraging performance for other tasks such as object detection, segmentation, depth estimation, and action recognition, etc. Girshick et al.~\cite{girshick2014rich} proposed R-CNN model (i.e., Regions with CNN features) for object detection. The `You Only Look Once (YOLO)' model was proposed by Redmon et al.~\cite{redmon2016you} for a unified, real-time object detection. Repala et al.~\cite{repala2018dual} built a dual CNN based unsupervised model for depth estimation. 
Recently, Singh et al.~\cite{singh2018recognizing} proposed Long Short-Term Memory (LSTM) networks and CNN based classifier to classify human actions.

The deep learning has been also utilized extensively for medical image and video analysis due to its capabilities to deal with complex data. In $2016$, IEEE Transactions on Medical Imaging published a special issue on deep learning in medical imaging which focused on the achievement of CNN and other deep learning based approaches~\cite{greenspan2016guest}. Litjens et al.~\cite{litjens2017survey} conducted a survey on deep learning in medical imaging by considering nearly 300 latest contributions, including image classification, object detection and segmentation tasks where deep learning techniques were used.  Esteva et al.~\cite{esteva2017dermatologist} proposed a deep CNN based classifier for skin cancer detection by training the model over a dataset of $1,29,450$ clinical images covering over $2,032$ different types of diseases. In $2017$, Rajpurkar et al.~\cite{rajpurkar2017chexnet} proposed CheXNet which is a $121$ layer CNN model. The ChexNet model is trained over Chest X-ray14 dataset which is one of the largest publicly available chest X-ray dataset containing $1,00,000$ X-ray images belonging to $14$ different diseases.

Xu et al.~\cite{xu2016stacked} proposed an unsupervised deep learning model called auto-encoder to classify cell nuclei, where the higher level features are classified using soft-max classifier. Korbar et al.~\cite{korbar2017looking}  introduced a deep neural network model to classify different types of colorectal polyps in whole-slide images. Very recently, Bychkov et al.~\cite{bychkov2018deep} proposed a classifier by combining the convolutional and recurrent neural network architectures for Colorectal cancer classification. 

Sirinukunwattana et al.~\cite{sirinukunwattana2016locality} proposed a convolutional neural network named as softmaxCNN$\_$IN27  to classify cell nuclei in histology images. Their softmaxCNN$\_$IN27 architecture has 5 trainable layers and $8,99,200$ learnable parameters. We have experimentally observed that the softmaxCNN$\_$IN27  model used by Sirinukunwattana et al.~\cite{sirinukunwattana2016locality} is not  deep enough as compared to the complexity of the histology image dataset. To overcome this problem, we have proposed a deep CNN model named as RCCNet having $7$ trainable layers with $1,512,868$ learnable parameters which outperforms softmaxCNN$\_$IN27~\cite{sirinukunwattana2016locality} for the histological routine colon cancer nuclei classification task.
 
The main objective of this paper is to develop an efficient and simple CNN architecture suitable for the classification of histological colon cancer images. The simplicity considered is in terms of the number of layers and number of trainable parameters, which are compared against the widely used CNN models such as AlexNet, CIFAR-VGG, GoogLeNet and WRN. In this work, we figured out that a careful consideration of number of trainable layers and trainable parameters can lead to an efficient CNN model. The proposed model is called the RCCNet which is used for the RCC classification task. Experimentally, we compared the proposed method with other popular models such as softmaxCNN$\_$IN27~\cite{sirinukunwattana2016locality}, softmaxCNN~\cite{sirinukunwattana2016locality}, AlexNet~\cite{krizhevsky2012imagenet}, CIFAR-VGG~\cite{liu2015very}, GoogLeNet~\cite{szegedy2015going}, and WRN~\cite{zagoruyko2016wide}. A promising performance is observed using the RCCNet in terms of the efficiency and accuracy.

The rest of the paper is organized as follows. Section \ref{pm} is devoted to the detailed description of the proposed RCCNet architecture. Section \ref{experimental_setup} presents the experimental setup including dataset description along with a description of compared methods. Results and Analysis are reported in section \ref{results_analysis}. Finally, section \ref{con} concludes the paper.

\section{Proposed RCCNet Architecture}
\label{pm}

Categorization of histology images is hard problem due to the high inter-class similarity and intra-class variablility.  The primary objective of our work is to design a Convolutional Neural Network (CNN) based architecture which classifies the colon cancer images. This section describes the proposed RCCNet which has seven trainable layers.

The proposed RCCNet architecture is illustrated in Fig. \ref{model}. In the proposed architecture, we considered histology images of dimension $32\times32\times3$ as input to the network. This CNN model has three blocks with seven trainable layers. In the $1^{st}$ block, two convolutional layers, viz., $Conv1$ and $Conv2$ are used just after the input layer. The $Conv2$ layer is followed by a pooling layer ($Pool1$) to reduce the spatial dimension by half. In the $2^{nd}$ block, two convolutional layers (i.e., $Conv3$ and $Conv4$ layers) are followed by another pooling layer ($Pool2$). In the $3^{rd}$ block, three fully connected layers, namely $FC1$, $FC2$, and $FC3$ are used in the proposed architecture. The input to $1^{st}$ layer of $3^{rd}$ block is basically the flattened features obtained from $Pool2$ layer. The $1^{st}$ convolutional layer $Conv1$ produces a $32\times32\times32$ dimensional feature map by convolving $32$ filters of dimension $3\times3\times3$. The zero padding by $1$ pixel in each direction is done in $Conv1$ layer to retain the same spatial dimensional feature map. The $Conv2$ layer has the $32$ filters of dimension $3\times3\times32$ with no padding which produces a $30\times30\times32$ dimensional feature map. The stride is set to $1$ in both $Conv1$ and $Conv2$ layers. In $Pool1$ layer, the sub-sampling with the receptive field of $2\times2$ is applied with a stride of $2$ and without padding which results in feature map of size $15\times15\times32$. The $Conv3$ layer produces $64$ feature maps of spatial dimension $15\times15$ (i.e., spatial dimension is retained by applying zero padding with a factor of 1), which is obtained by applying $64$ filters of dimension $3\times3\times32$ with a stride of $1$. Similar to $Conv2$ layer, $Conv4$ layer also does not apply padding and uses stride of $1$. The $Conv4$ layer produces $64$ features maps of dimension $13\times13$, obtained by convolving the $64$ filters of size $3\times3\times64$. The second sub-sampling layer $Pool2$ also uses the kernel size of $2\times2$ with a stride of $2$, which results in a $6\times6\times64$ dimensional feature map. The right and bottom border feature values of input are not considered in $Pool2$ layer to get rid of dimension mismatch between input and kernel size. The feature map generated by $Pool2$ layer is flattened into a single feature vector of length $2304$ before $3^{rd}$ block (i.e., fully connected layers). So, the input to $FC1$ layer is $2304$ dimensional feature vector and output is $512$ dimensional feature vector. Both input and output to $FC2$ layer is $512$ dimensional feature vectors. The last fully connected layer $FC3$ takes the input of dimension $512$ (i.e., the output of $FC2$ layer) and produces the $4$ values as the output corresponding to the scores for $4$ classes. This architecture consists of $1,512,868$ trainable parameters from 7 trainable layers (i.e., $Conv1$, $Conv2$, $Conv3$, $Conv4$, $FC1$, $FC2$, and $FC3$ layers). 

On top of the last fully connected layer $FC3$ of proposed RCCNet model, a `softmax classifier' for multi-class classification is used to generate the probabilities for each class. The probabilities generated by the `softmax classifier' is further used to compute the loss during training phase and to find the predicted class during testing phase. 

\subsection{Training Phase}
The categorical cross entropy loss is computed during the training phase. The parameters (weights) of the network are updated by finding the gradient of parameters with respect to the loss function. The cross-entropy loss (also known as the log loss) is used to compute the performance of a classifier whose output is a probability value ranging between $0$ and $1$.
Let $x$ be a three-dimensional input image to the network with class label $c \in C$ where $ C = \{c_1, c_2, ... c_n\}$ is the set of class labels. In the current classification task, $C = \{1,2,3,4\}$. The output of the network is a vector $y$ which is,
\begin{equation}
y=f(x)
\end{equation}
where $f$ denotes the forward pass computation function and $y=[y_1,y_2,...,y_{c_n}]$ represents the class scores for the $n$ classes.
The cross-entropy loss for $x$, assuming that the target class (as given in the training set) is $c_i$, 

\begin{equation} \label{entropy}
L = -log\left( \frac{e^{y_i}}{\sum_{k=c_1}^{c_n}{e^{y_k}}}\right).
\end{equation}

The total loss over a mini-batch of training examples is considered in the training process.

\subsection{Testing Phase}
At test time, for a given input image, the class label having the highest score is the predicted class label. The predicted class label $c_j$ is computed as,
\begin{equation}
c_j = \displaystyle\arg \max_i p(y_i)
\end{equation}
where $p(y_i)$ is the probability that $x$ belongs to class $c_i$, which is computed as follows.
\begin{equation} \label{softmax}
p(y_i) = \frac{e^{y_i}}{\sum_{k=c_1}^{c_n}{e^{y_k}}}.
\end{equation}

\addtolength{\textheight}{-3cm}   

\section{Experimental Setup} \label{experimental_setup}
This section is devoted to present the experimental setting including dataset description, a briefing about the compared models, training details and the evaluation criteria.  

\subsection{Dataset Description}
\begin{figure}[t]
\begin{center}
\vspace*{0.2cm} \includegraphics[width=\linewidth, height = 4.5 cm]{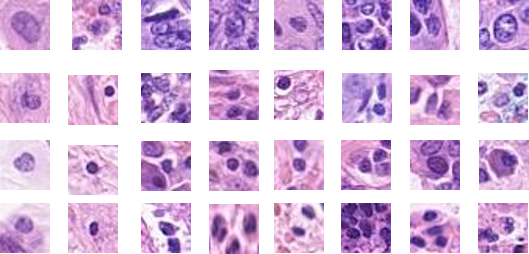}
\setlength{\belowcaptionskip}{-15pt}
\caption{The sample patches from each class of ``CRCHistoPhenotypes" dataset~\cite{sirinukunwattana2016locality}. The $1^{st}$, $2^{nd}$, $3^{rd}$, and $4^{th}$ rows show the patches from the classes, `Epithelial', `Fibroblast', `Inflammatory', and `Miscellaneous', respectively. The  different columns in a row represent different samples from the same class.}

\label{fig:dataset}
\end{center}
\end{figure}

In order to find the performance of the proposed RCCNet for the task, we have used a publicly available `CRCHistoPhenotypes' dataset\footnote{\url{https://warwick.ac.uk/fac/sci/dcs/research/tia/data/crchistolabelednucleihe}} which consists of the histological routine colon cancer nuclei patches~\cite{sirinukunwattana2016locality}. This dataset consists of $22444$ nuclei patches that belong to the four classes, namely, `Epithelial', `Inflammatory', `Fibroblast', and `Miscellaneous'. In total, there are $7722$ patches from the `Epithelial' class, $5712$ patches from the `Fibroblast' class, $6971$ patches from the `Inflammatory' class and the remaining $2039$ patches from the `Miscellaneous' class. The dimension of each patch is $32\times32\times3$. The sample cell nuclei patches from the `CRCHistoPhenotypes' dataset is given in Fig. \ref{fig:dataset}.

\subsection{Compared CNN Models}
\label{comparison}
In order to justify the performance of the proposed RCCNet for the task, five state-of-the-art CNN models are implemented and a comparison is drawn. 
A brief overview of these architectures is given in the rest of this subsection. 

\subsubsection{softmaxCNN\_IN27~\cite{sirinukunwattana2016locality}}
\label{softmaxCNN}
Sirinukunwattana et al.~\cite{sirinukunwattana2016locality} proposed softmaxCNN$\_$IN27 architecture for the classification task. This model has $5$ learnable layers including $2$ convolutional and $3$ fully connected layers. Each convolutional layer is followed by a max-pooling layer to reduce the spatial dimension by half. The $1^{st}$ convolutional layer has $36$ filters of size $4\times4$, which results in a feature map of dimension $24\times24\times36$. The $1^{st}$ max-pool reduces the dimension of feature map to $12\times12\times36$. The $2^{nd}$ convolutional layer has $48$ filters of kernel size $3\times3$, which produces a feature map of dimension $10\times10\times48$. The $2^{nd}$ max-pooling produces the $5\times5\times48$ dimensional feature map. It is further followed by three fully connected layers, which have $512$, $512$, and $4$ nodes, respectively. softmaxCNN$\_$IN27 model consists of $899200$ trainable parameters.

We also modified the architecture of sirinukunwattana et al.~\cite{sirinukunwattana2016locality} with some minimal changes to make it suitable for $32\times32\times3$ dimensional input. It is called  softmaxCNN. Initially, the input images are up-sampled to $33\times33\times3$. Then, zero padding by $1$ pixel in each direction is done, which results in a $35\times35\times3$ dimensional image. The $1^{st}$ convolution layer produces a $31\times31\times32$ dimensional feature map by convolving $32$ filters of size $5\times5\times3$. This is followed by another convolution layer, which produces a $27\times27\times32$ dimensional feature map by applying $32$ filters of size $5\times5\times32$.  The rest of the architecture is same as the original softmaxCNN$\_$IN27~\cite{sirinukunwattana2016locality}. softmaxCNN model consists of $944032$ trainable parameters from $7$ trainable layers.

\subsubsection{AlexNet~\cite{krizhevsky2012imagenet}}
\label{AlexNet}
AlexNet~\cite{krizhevsky2012imagenet} is the most popular CNN architecture, originally proposed for natural image classification. Initially we tried to make use of the original AlexNet~\cite{krizhevsky2012imagenet} architecture by up-sampling the image dimension from $27\times27\times3$ to $227\times227\times3$. However, we experimentally observed no improvement even after training this model for 200 epochs. With this observation, we made minimal modifications to the AlexNet to fit for low resolution images. The image dimensions are up-sampled from $32\times32\times3$ to $33\times33\times3$. Then, zero padding by $1$ pixel in each direction is done, which results in a $35\times35\times3$ dimensional image. This is followed by $1^{st}$ convolution layer which produces a $31\times31\times96$ dimensional feature vector by applying $96$ filters of dimension $5\times5\times3$.  The $2^{nd}$ convolutional layer produces the feature map of dimension $27\times27\times256$ by convolving $256$ filters of dimension $5\times5\times96$. The rest of the architecture is same as original AlexNet~\cite{krizhevsky2012imagenet} (the last fully connected layer is modified to have $4$ neurons instead of $1000$). This architecture corresponds to $197,731,396$ trainable parameters with 8 trainable layers.

\subsubsection{CIFAR-VGG ~\cite{simonyan2014very, liu2015very}} \label{VGG16}
Originally, the VGG-16 model was introduced by Simonyan et al.~\cite{simonyan2014very} for ImageNet challenge. Liu et al.~\cite{liu2015very} proposed a modified VGG-16 architecture (CIFAR-VGG) for training low scale images like CIFAR-10~\cite{krizhevsky2009learning}. We have utilized the CIFAR-VGG architecture~\cite{liu2015very} to train over histology images by changing the number of neurons in last FC layer to $4$. This model has $16$ trainable layers with $8,956,484$ trainable parameters.

\subsubsection{GoogLeNet~\cite{szegedy2015going}} \label{googlenet}
GoogLeNet~\cite{szegedy2015going} is the winner of ILSVRC 2014, which consists 22 learnable layers. GoogLeNet~\cite{szegedy2015going} is originally proposed for classification of large scale natural images. We made minimal changes to the GoogLeNet architecture~\cite{szegedy2015going} to work for low-resolution images. The $1^{st}$ convolutional layer produces a $30\times30\times64$ dimensional feature map by applying $64$ filters of dimension $3\times3\times3$. Then, $2^{nd}$ convolution layer computes a $28\times28\times128$ dimensional feature map by applying $128$ filters of dimension $3\times3\times64$. This is followed by an inception block, which results in a $28\times28\times256$ dimensional feature vector. The remaining part of the model is similar to original GoogLeNet~\cite{szegedy2015going} except the last fully connected layer which is is modified to have $4$ neurons instead of $1000$. This CNN model corresponds to $11,526,988$ trainable parameters.

\begin{table*}
  \begin{center}
  \vspace*{0.2cm} 
   \caption{The performance comparison of proposed RCCNet with state-of-the-art softmaxCNN$\_$IN27, AlexNet\cite{krizhevsky2012imagenet}, CIFAR-VGG\cite{liu2015very}, GoogLeNet\cite{szegedy2015going}, and WRN\cite{zagoruyko2016wide} models in terms of the Training Accuracy, Testing Accuracy, Overfitting $\%$, Training F1 Score, and Testing F1 Score. The dimension of input images to softmaxCNN$\_$IN27 model~\cite{sirinukunwattana2016locality} is $27\times27\times3$, whereas for other models input size is $32\times32\times3$. The best accuracies and F1 scores are highlighted in \textbf{bold}.}
   
    \scalebox{0.9}{
      \begin{tabular}{| c | c | c | c c c | c c | c |} \hline
      \multirow{2}{*}{\textbf{Model Name}} &  \multirow{2}{*}{\parbox{1.5cm}{\textbf{$\#$Trainable Parameters}}} &  \multirow{2}{*}{\parbox{1.8cm}{\textbf{Training Time (in minutes)}}} & \multicolumn{3}{ c |}{\textbf{Classification Accuracy}} & \multicolumn{2}{ c |}{\textbf{Weighted Average F1 Score}}  \\ 
\cline{4-8}
&  & & Training Accuracy \%& Testing Accuracy \%  & Overfitting $\%$ & Training F1 score & Testing F1 score\\ \hline

softmaxCNN$\_$IN27~\cite{sirinukunwattana2016locality}  &  899,200 & 27.673 & 82.90  & 71.15  & 11.75 & 0.9332 &  0.7124\\

softmaxCNN~\cite{sirinukunwattana2016locality} &  944,032 & 28.774 & 83.63  & 73.71  & 9.92 & 0.9581 &  0.7439\\

AlexNet~\cite{krizhevsky2012imagenet}& 197,731,396 & 818.675 & 92.38 & 76.96 & 15.42 & 0.9897 &  0.7664\\

CIFAR-VGG~\cite{liu2015very}  & 8,956,484 & 102.259 &85.43  & 75.94  & 9.49 & 0.9983 &  0.7708\\

GoogLeNet~\cite{szegedy2015going}  & 11,526,988 & 391.566 &\textbf{99.99 } & 78.99  & 21.00 & \textbf{0.9999} & 0.7856\\

WRN~\cite{zagoruyko2016wide}  & 23,542,788 & 77.829 & 98.61  & 61.88  & 36.73 &  0.9968 &  0.6227\\ \hline

RCCNet (Proposed) & 1,512,868 & 31.404 &  89.79 &   \textbf{80.61}  & \textbf{9.18} & 0.9037 & \textbf{0.7887}\\ \hline

      \end{tabular}
    }
    \label{accuracy_NoDA}
    
   \end{center}
   
\end{table*}

\subsubsection{WRN~\cite{zagoruyko2016wide}}
\label{resnet}
He et al.~\cite{he2016deep} introduced the concept of residual networks for natural image classification. Zagoruyko et al.~\cite{zagoruyko2016wide} proposed a wide Residual Network (WRN) to train low resolution images of CIFAR-10 dataset. In this paper, we have adapted the WRN architecture~\cite{zagoruyko2016wide} for comparison purpose. The number of nodes in the last fully connected layer is changed to $4$ corresponding to the number of classes in used histology dataset. The WRN architecture used in this paper consists of $23,542,788$ trainable parameters.

\subsection{Training Details}
The initial value of the learning rate is considered as $6\times10^{-5}$, and iteratively decreased with a factor of $\sqrt[2]{0.1}$ if there is no improvement in validation loss during training. The rectified linear unit $(ReLU)$~\cite{krizhevsky2012imagenet} is employed as the activation function in all the implemented models. To reduce over fitting, dropout ~\cite{srivastava2014dropout} is used after $ReLU$ of each fully connected layer with a rate of $0.5$ and Batch Normalization \cite{ioffe2015batch} used after every trainable layer(except last $FC$ layer) after $ReLU$ is applied. All the models are trained for $500$ epochs using Adam optimizer~\cite{kingma2014adam} with $\beta_1 = 0.9$, $\beta_2 = 0.99$, and $decay = 1\times10^{-6}$. The $80\%$ of entire dataset ($i.e., 17,955$ images) is used for the training and remaining $20\%$ ($i.e., 4,489$ images) is used to test the performance.

\subsection{Evaluation Criteria}
In order to assess the performance of CNN models, we have considered two performance measures accuracy and weighted average F1 score. In this paper, the training time is also considered as one of the evaluation metrics to judge the efficiency of the CNN models.

\begin{figure}
\includegraphics[width=\linewidth, height=4.5 cm]{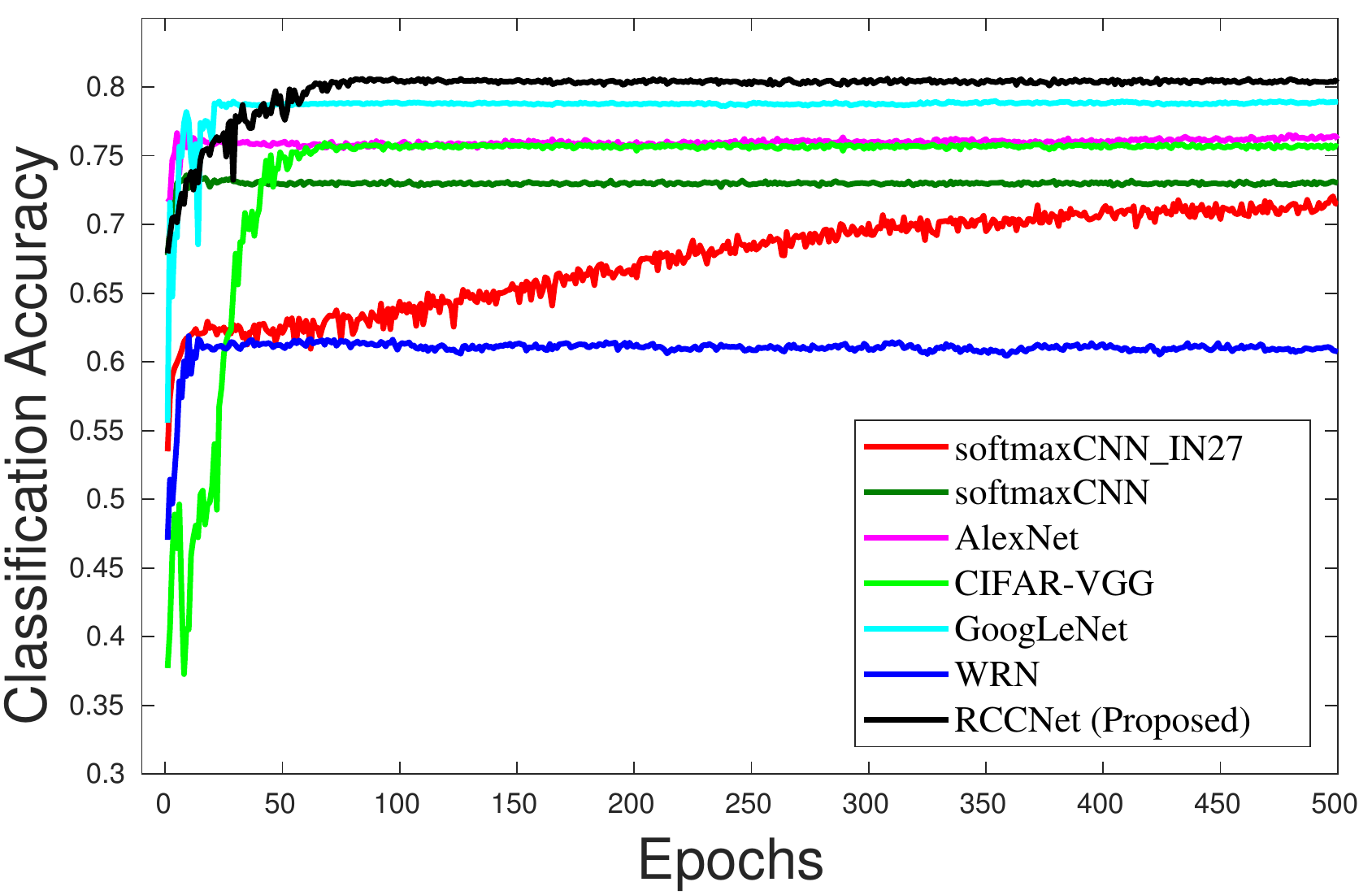}
\setlength{\belowcaptionskip}{-15pt}
\caption{The Epoch wise test accuracies for different CNN models used in this paper for experiments.}
\label{testing_accuracies_NoDA_label}
\end{figure}

\section{Results and Analysis}
\label{results_analysis}

We have conducted the extensive experiments to compare the performance of proposed RCCNet model with other state-of-the-art CNN models like softmaxCNN~\cite{sirinukunwattana2016locality}, AlexNet~\cite{krizhevsky2012imagenet}, CIFAR-VGG~\cite{liu2015very}, GoogLeNet~\cite{szegedy2015going}, and WRN~\cite{zagoruyko2016wide}. Table \ref{accuracy_NoDA} presents the performance comparison among the CNN models in terms of the number of trainable parameters, training time, training accuracy, testing accuracy, amount of over-fitting, training F1 score, and testing F1 score. Followings are the main observations from the results of Table \ref{accuracy_NoDA}:
\begin{enumerate}

\item The proposed RCCNet model outperforms the other CNN models both in terms of test accuracy and test weighted F1 score because the proposed model is highly optimized for histological routine colon cancer images.
\item The proposed RCCNet model has the lowest number of trainable parameters except softmaxCNN\_IN27 and softmaxCNN~\cite{sirinukunwattana2016locality}. Whereas, the AlexNet~\cite{krizhevsky2012imagenet} model has largest number of learnable parameters.
\item The softmaxCNN model~\cite{sirinukunwattana2016locality} proposed originally for histological routine colon cancer images is not enough complex, whereas our model is enough complex to produce a reasonable performance.
\item The proposed RCCNet model is better generalized as compared to other CNN models and results in lowest amount of over-fitting as depicted in Table \ref{accuracy_NoDA}. The highest amount of over-fitting is observed for wide residual network (WRN)~\cite{zagoruyko2016wide}. This analysis points out that the amount of over-fitting is closely related to the network structure like depth of network, number of learnable parameters, and type of network (i.e., plain/inception/residual). 
\item The test accuracy using RCCNet architecture is improved by $13.3 \%$, $9.36 \%$, $4.74$ \%, $2.05 \%$, and $30.4\%$ as compared to softmaxCNN$\_$IN27~\cite{sirinukunwattana2016locality}, softmaxCNN~\cite{sirinukunwattana2016locality},  AlexNet~\cite{krizhevsky2012imagenet}, GoogLeNet~\cite{szegedy2015going}, and WRN~\cite{zagoruyko2016wide} models, respectively due to the even size filters in softmaxCNN$\_$IN27 and softmaxCNN~\cite{sirinukunwattana2016locality} and high over-fitting occurred in other models.
\item The proposed RCCNet is more efficient in terms of the training time which is improved by $96.16\%$, $69.28\%$, $91.97\%$, and $56.64\%$ as compared to AlexNet~\cite{krizhevsky2012imagenet}, CIFAR-VGG~\cite{liu2015very}, GoogLeNet~\cite{szegedy2015going}, and WRN~\cite{zagoruyko2016wide}, respectively due to the less number of learnable parameters and plain network architecture.

\end{enumerate}

Fig. \ref{testing_accuracies_NoDA_label} shows the comparison among test accuracies of implemented CNN architectures. From Fig. \ref{testing_accuracies_NoDA_label}, it is observed that
the AlexNet~\cite{krizhevsky2012imagenet}, GoogLeNet~\cite{szegedy2015going} and WRN~\cite{zagoruyko2016wide}
converge quickly compared to other CNN architectures. The softmaxCNN$\_$IN27~\cite{sirinukunwattana2016locality} model is slow in terms of the convergence. However, the proposed RCCNet architecture is very reasonable and converges smoothly.

\section{Conclusion} \label{con}
In this paper, we have proposed an efficient convolutional neural network based classification model to classify colon cancer images. The proposed RCCNet model is highly compact and optimized for histological low-resolution patches. Only $7$ plain trainable layers are used with $1,512,868$ trainable parameters. The classification experiments are performed over histological routine colon cancer patches. The performance of the proposed RCCNet model is compared with the other popular models like AlexNet, CIFAR-VGG, GoogLeNet, and WRN. The experimental results point out that the RCCNet is better generalizes and outperforms other models in terms of the test accuracy and weighted average F1 score. The proposed RCCNet model attains $80.61\%$ classification accuracy and $0.7887$ weighted average F1 score. The RCCNet is also highly efficient in terms of the training time as compared to deeper and complex networks.

\section*{ACKNOWLEDGMENTS}
This research is supported in part by Science and Engineering Research Board (SERB), Govt. of India, Grant No. ECR/2017/000082.

\bibliographystyle{IEEEtran}  
\bibliography{egbib} 

\begin{thebibliography}{10}
\providecommand{\url}[1]{#1}
\csname url@samestyle\endcsname
\providecommand{\newblock}{\relax}
\providecommand{\bibinfo}[2]{#2}
\providecommand{\BIBentrySTDinterwordspacing}{\spaceskip=0pt\relax}
\providecommand{\BIBentryALTinterwordstretchfactor}{4}
\providecommand{\BIBentryALTinterwordspacing}{\spaceskip=\fontdimen2\font plus
\BIBentryALTinterwordstretchfactor\fontdimen3\font minus
  \fontdimen4\font\relax}
\providecommand{\BIBforeignlanguage}[2]{{%
\expandafter\ifx\csname l@#1\endcsname\relax
\typeout{** WARNING: IEEEtran.bst: No hyphenation pattern has been}%
\typeout{** loaded for the language `#1'. Using the pattern for}%
\typeout{** the default language instead.}%
\else
\language=\csname l@#1\endcsname
\fi
#2}}
\providecommand{\BIBdecl}{\relax}
\BIBdecl

\bibitem{sirinukunwattana2016locality}
K.~Sirinukunwattana, S.~E.~A. Raza, Y.-W. Tsang, D.~R. Snead, I.~A. Cree, and
  N.~M. Rajpoot, ``Locality sensitive deep learning for detection and
  classification of nuclei in routine colon cancer histology images,''
  \emph{IEEE transactions on medical imaging}, vol.~35, no.~5, pp. 1196--1206,
  2016.

\bibitem{siegel2017colorectal}
R.~L. Siegel, K.~D. Miller, S.~A. Fedewa, D.~J. Ahnen, R.~G. Meester, A.~Barzi,
  and A.~Jemal, ``Colorectal cancer statistics, 2017,'' \emph{CA: a cancer
  journal for clinicians}, vol.~67, no.~3, pp. 177--193, 2017.

\bibitem{veta2014breast}
M.~Veta, J.~P. Pluim, P.~J. Van~Diest, and M.~A. Viergever, ``Breast cancer
  histopathology image analysis: A review,'' \emph{IEEE Transactions on
  Biomedical Engineering}, vol.~61, no.~5, pp. 1400--1411, 2014.

\bibitem{arif2007classification}
M.~Arif and N.~Rajpoot, ``Classification of potential nuclei in prostate
  histology images using shape manifold learning,'' in \emph{Machine Vision,
  2007. ICMV 2007. International Conference on}.\hskip 1em plus 0.5em minus
  0.4em\relax IEEE, 2007, pp. 113--118.

\bibitem{sharma2015multi}
H.~Sharma, N.~Zerbe, D.~Heim, S.~Wienert, H.-M. Behrens, O.~Hellwich, and
  P.~Hufnagl, ``A multi-resolution approach for combining visual information
  using nuclei segmentation and classification in histopathological images.''
  in \emph{VISAPP (3)}, 2015, pp. 37--46.

\bibitem{jones2009scoring}
T.~R. Jones, A.~E. Carpenter, M.~R. Lamprecht, J.~Moffat, S.~J. Silver, J.~K.
  Grenier, A.~B. Castoreno, U.~S. Eggert, D.~E. Root, P.~Golland \emph{et~al.},
  ``Scoring diverse cellular morphologies in image-based screens with iterative
  feedback and machine learning,'' \emph{Proceedings of the National Academy of
  Sciences}, vol. 106, no.~6, pp. 1826--1831, 2009.

\bibitem{chang2013classification}
H.~Chang, A.~Borowsky, P.~Spellman, and B.~Parvin, ``Classification of tumor
  histology via morphometric context,'' in \emph{Proceedings of the IEEE
  Conference on Computer Vision and Pattern Recognition}, 2013, pp. 2203--2210.

\bibitem{lecun2015deep}
Y.~LeCun, Y.~Bengio, and G.~Hinton, ``Deep learning,'' \emph{nature}, vol. 521,
  no. 7553, p. 436, 2015.

\bibitem{krizhevsky2012imagenet}
A.~Krizhevsky, I.~Sutskever, and G.~E. Hinton, ``Imagenet classification with
  deep convolutional neural networks,'' in \emph{Advances in neural information
  processing systems}, 2012, pp. 1097--1105.

\bibitem{simonyan2014very}
K.~Simonyan and A.~Zisserman, ``Very deep convolutional networks for
  large-scale image recognition,'' \emph{arXiv preprint arXiv:1409.1556}, 2014.

\bibitem{szegedy2015going}
C.~Szegedy, W.~Liu, Y.~Jia, P.~Sermanet, S.~Reed, D.~Anguelov, D.~Erhan,
  V.~Vanhoucke, and A.~Rabinovich, ``Going deeper with convolutions,'' in
  \emph{Proceedings of the IEEE conference on computer vision and pattern
  recognition}, 2015, pp. 1--9.

\bibitem{he2016deep}
K.~He, X.~Zhang, S.~Ren, and J.~Sun, ``Deep residual learning for image
  recognition,'' in \emph{Proceedings of the IEEE conference on computer vision
  and pattern recognition}, 2016, pp. 770--778.

\bibitem{girshick2014rich}
R.~Girshick, J.~Donahue, T.~Darrell, and J.~Malik, ``Rich feature hierarchies
  for accurate object detection and semantic segmentation,'' in
  \emph{Proceedings of the IEEE conference on computer vision and pattern
  recognition}, 2014, pp. 580--587.

\bibitem{redmon2016you}
J.~Redmon, S.~Divvala, R.~Girshick, and A.~Farhadi, ``You only look once:
  Unified, real-time object detection,'' in \emph{Proceedings of the IEEE
  conference on computer vision and pattern recognition}, 2016, pp. 779--788.

\bibitem{repala2018dual}
V.~K. Repala and S.~R. Dubey, ``Dual cnn models for unsupervised monocular
  depth estimation,'' \emph{arXiv preprint arXiv:1804.06324}, 2018.

\bibitem{singh2018recognizing}
K.~K. Singh and S.~Mukherjee, ``Recognizing human activities in videos using
  improved dense trajectories over lstm,'' in \emph{Computer Vision, Pattern
  Recognition, Image Processing, and Graphics: 6th National Conference,
  NCVPRIPG 2017, Mandi, India, December 16-19, 2017, Revised Selected Papers
  6}.\hskip 1em plus 0.5em minus 0.4em\relax Springer, 2018, pp. 78--88.

\bibitem{greenspan2016guest}
H.~Greenspan, B.~van Ginneken, and R.~M. Summers, ``Guest editorial deep
  learning in medical imaging: Overview and future promise of an exciting new
  technique,'' \emph{IEEE Transactions on Medical Imaging}, vol.~35, no.~5, pp.
  1153--1159, 2016.

\bibitem{litjens2017survey}
G.~Litjens, T.~Kooi, B.~E. Bejnordi, A.~A.~A. Setio, F.~Ciompi, M.~Ghafoorian,
  J.~A. van~der Laak, B.~van Ginneken, and C.~I. S{\'a}nchez, ``A survey on
  deep learning in medical image analysis,'' \emph{Medical image analysis},
  vol.~42, pp. 60--88, 2017.

\bibitem{esteva2017dermatologist}
A.~Esteva, B.~Kuprel, R.~A. Novoa, J.~Ko, S.~M. Swetter, H.~M. Blau, and
  S.~Thrun, ``Dermatologist-level classification of skin cancer with deep
  neural networks,'' \emph{Nature}, vol. 542, no. 7639, p. 115, 2017.

\bibitem{rajpurkar2017chexnet}
P.~Rajpurkar, J.~Irvin, K.~Zhu, B.~Yang, H.~Mehta, T.~Duan, D.~Ding, A.~Bagul,
  C.~Langlotz, K.~Shpanskaya \emph{et~al.}, ``Chexnet: Radiologist-level
  pneumonia detection on chest x-rays with deep learning,'' \emph{arXiv
  preprint arXiv:1711.05225}, 2017.

\bibitem{xu2016stacked}
J.~Xu, L.~Xiang, Q.~Liu, H.~Gilmore, J.~Wu, J.~Tang, and A.~Madabhushi,
  ``Stacked sparse autoencoder (ssae) for nuclei detection on breast cancer
  histopathology images,'' \emph{IEEE transactions on medical imaging},
  vol.~35, no.~1, pp. 119--130, 2016.

\bibitem{korbar2017looking}
B.~Korbar, A.~M. Olofson, A.~P. Miraflor, C.~M. Nicka, M.~A. Suriawinata,
  L.~Torresani, A.~A. Suriawinata, and S.~Hassanpour, ``Looking under the hood:
  Deep neural network visualization to interpret whole-slide image analysis
  outcomes for colorectal polyps,'' in \emph{Computer Vision and Pattern
  Recognition Workshops (CVPRW), 2017 IEEE Conference on}.\hskip 1em plus 0.5em
  minus 0.4em\relax IEEE, 2017, pp. 821--827.

\bibitem{bychkov2018deep}
D.~Bychkov, N.~Linder, R.~Turkki, S.~Nordling, P.~E. Kovanen, C.~Verrill,
  M.~Walliander, M.~Lundin, C.~Haglund, and J.~Lundin, ``Deep learning based
  tissue analysis predicts outcome in colorectal cancer,'' \emph{Scientific
  reports}, vol.~8, no.~1, p. 3395, 2018.

\bibitem{liu2015very}
S.~Liu and W.~Deng, ``Very deep convolutional neural network based image
  classification using small training sample size,'' in \emph{Pattern
  Recognition (ACPR), 2015 3rd IAPR Asian Conference on}.\hskip 1em plus 0.5em
  minus 0.4em\relax IEEE, 2015, pp. 730--734.

\bibitem{zagoruyko2016wide}
S.~Zagoruyko and N.~Komodakis, ``Wide residual networks,'' \emph{arXiv preprint
  arXiv:1605.07146}, 2016.

\bibitem{krizhevsky2009learning}
A.~Krizhevsky and G.~Hinton, ``Learning multiple layers of features from tiny
  images,'' 2009.

\bibitem{srivastava2014dropout}
N.~Srivastava, G.~Hinton, A.~Krizhevsky, I.~Sutskever, and R.~Salakhutdinov,
  ``Dropout: A simple way to prevent neural networks from overfitting,''
  \emph{The Journal of Machine Learning Research}, vol.~15, no.~1, pp.
  1929--1958, 2014.

\bibitem{ioffe2015batch}
S.~Ioffe and C.~Szegedy, ``Batch normalization: Accelerating deep network
  training by reducing internal covariate shift,'' \emph{arXiv preprint
  arXiv:1502.03167}, 2015.

\bibitem{kingma2014adam}
D.~P. Kingma and J.~Ba, ``Adam: A method for stochastic optimization,''
  \emph{arXiv preprint arXiv:1412.6980}, 2014.

\end{thebibliography}
\end{document}